\documentclass[11pt,a4paper]{article}
\usepackage[hyperref]{acl2020}
\usepackage{times}
\usepackage{latexsym}

\usepackage{microtype}

\usepackage{multirow}
\usepackage{bm}

\usepackage{enumitem}
\usepackage{amsfonts}
\usepackage{amssymb}

\usepackage{amsmath}
\usepackage{graphicx}
\usepackage{color}
\usepackage{subcaption}
\definecolor{newblue}{RGB}{70,118,214}
\definecolor{neworange}{RGB}{206,151,90}
\definecolor{newpurple}{RGB}{131,106,212}
\definecolor{newgreen}{RGB}{58,105,57}
\definecolor{newpink}{RGB}{240,128,128}

\definecolor{Orange}{rgb}{1,0.5,0}

\aclfinalcopy 
\def\aclpaperid{2986} 

\newcommand*{\affmark}[1][*]{\textsuperscript{#1}}

\title{Parallel Data Augmentation for Formality Style Transfer}

\author{Yi Zhang\affmark[1]\thanks{ ~~Work done during the internship at Microsoft Research.}, \
        Tao Ge\affmark[2],\ 
        Xu Sun\affmark[1] \\
  \affmark[1]MOE Key Lab of Computational Linguistics, School of EECS, Peking University  \\
  \affmark[2]Microsoft Research Asia, Beijing, China  \\
  \texttt{\{zhangyi16,xusun\}@pku.edu.cn} \\
  \texttt{tage@microsoft.com} \\}

\date{}

\begin{document}
\maketitle
\begin{abstract}
The main barrier to progress in the task of Formality Style Transfer is the inadequacy of training data. In this paper, we study how to augment parallel data and propose novel and simple data augmentation methods for this task to obtain useful sentence pairs with easily accessible models and systems. Experiments demonstrate that our augmented parallel data largely helps improve formality style transfer when it is used to pre-train the model, leading to the state-of-the-art results in the GYAFC benchmark dataset\footnote{Our augmented data is available at \url{https://github.com/lancopku/Augmented_Data_for_FST}}.

\end{abstract}

\section{Introduction}\label{sec:intro}
Formality style transfer (FST) is defined as the task of automatically transforming a piece of text in one particular formality style into another \cite{DBLP:conf/naacl/RaoT18}. For example, given an informal sentence, FST aims to preserve the style-independent content and output a formal sentence. 

Previous work tends to leverage neural networks \cite{DBLP:journals/corr/abs-1903-06353,DBLP:conf/coling/NiuRC18,wang-etal-2019-harnessing} such as seq2seq models to address this challenge due to their powerful capability and large improvement over the traditional rule-based approaches \cite{DBLP:conf/naacl/RaoT18}. However, the performance of the neural network approaches is still limited by the inadequacy of training data: the public parallel corpus for FST training -- GYAFC \cite{DBLP:conf/naacl/RaoT18} -- contains only approximately 100K sentence pairs, which can hardly satiate the neural models with millions of parameters.

\begin{figure}[t]
    \centering
    \scalebox{1.0}{
    \includegraphics[width=7.5cm]{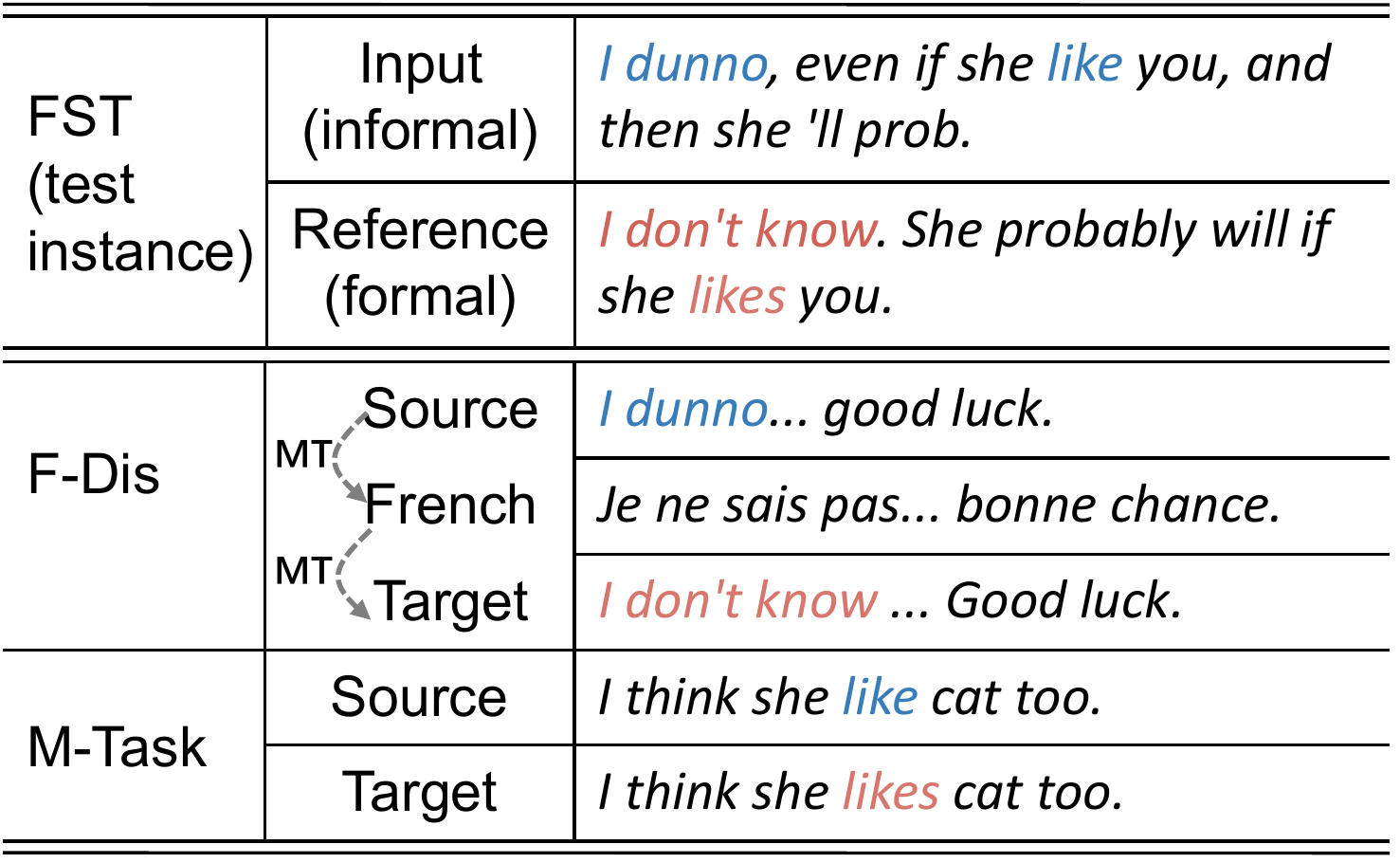}}
    \caption{An example that \textbf{F}ormality \textbf{S}tyle \textbf{T}ransfer (FST) benefits from data augmented via \textbf{f}ormality \textbf{dis}crimination (\textbf{F-Dis}) and \textbf{m}ulti-\textbf{task} transfer (\textbf{M-Task}). The mapping knowledge indicated by the color (\textcolor{newblue}{blue}$\to$\textcolor{newpink}{pink}) in FST test instance occur in the pairs augmented by F-Dis and M-Task. F-Dis identifies useful sentence pairs from paraphrased sentence pairs generated by cross-lingual MT, while M-Task utilizes training data from GEC to help formality improvement.}
    \label{example}
\end{figure}

To tackle the data sparsity problem for FST, we propose to augment parallel data with three specific data augmentation methods to help improve the model's generalization ability and reduce the overfitting risk. Besides applying the widely used back translation (BT) method~\citep{DBLP:conf/acl/SennrichHB16} in Machine Translation (MT) to FST, our data augmentation methods include formality discrimination (F-Dis) and multi-task transfer (M-Task). They are both novel and effective in generating parallel data that introduces additional formality transfer knowledge that cannot be derived from the original training data. Specifically, F-Dis identifies useful pairs from the paraphrased pairs generated by cross-lingual MT; while M-task leverages the training data of Grammatical Error Correction (GEC) task to improve formality, as shown in Figure \ref{example}.

Experimental results show that our proposed data augmentation methods can harvest large amounts of augmented parallel data for FST. The augmented parallel data proves helpful and significantly helps improve formality style transfer when it is used to pre-train the model, allowing the model to achieve the state-of-the-art results in the GYAFC benchmark dataset.





\section{Approach}

\subsection{Data Augmentation for Formality Style Transfer}\label{subsec:augmentation}
We study three data augmentation methods for formality style transfer: back translation, formality discrimination, and multi-task transfer. We focus on informal$\to$formal style transfer since it is more practical in real application scenarios.

\subsubsection{Back translation}\label{subsubsec:bt}
The original idea of back translation (BT) \cite{DBLP:conf/acl/SennrichHB16} is to train a target-to-source seq2seq~\citep{DBLP:conf/nips/SutskeverVL14,DBLP:journals/corr/ChoMGBSB14} model and use the model to generate source language sentences from target monolingual sentences, establishing synthetic parallel sentences. We generalize it as our basic data augmentation method and use the original parallel data to train a seq2seq model in the formal-to-informal direction. Then, we can feed formal sentences to this model that is supposed to be capable of generating their informal counterparts. The formal input and the informal output sentences can be paired to establish augmented parallel data.

\subsubsection{Formality discrimination}\label{subsubsec:f-dis}

\begin{figure}[t]
    \centering
    \includegraphics[width=7.5cm]{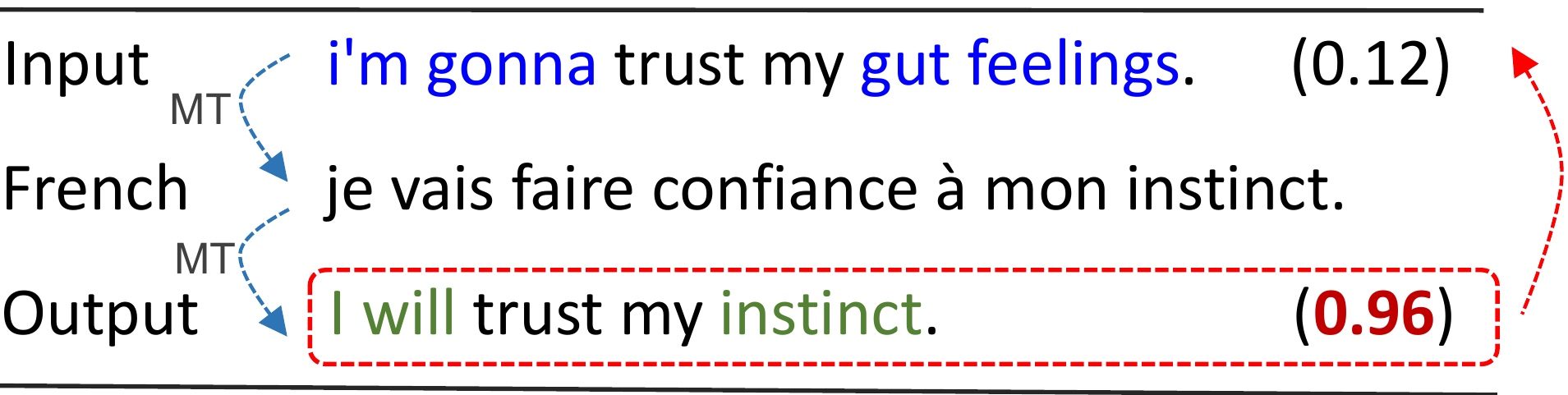}
    \caption{Formality discrimination for FST. The numbers following the sentences are formality scores predicted by a formality discriminator. The pair (connected by the red dashed arrow) that obtains significant formality improvement will be kept as augmented data.}
    \label{fig:fst_boost}
\end{figure}

According to the observation that an informal sentence tends to become a formal sentence after a round-trip translation by MT models that are mainly trained with formal text like news, we propose a novel method called formality discrimination to generate formal rewrites of informal source sentences by means of cross-lingual MT models. A typical example is shown in Figure \ref{fig:fst_boost}.

To this end, we collect a number of potentially informal English sentences (e.g., from online forums). Formally, we denote the collected sentences as $\mathcal{S}=\{\bm{s}_i\}_{i=1}^{|\mathcal{S}|}$ where $\bm{s}_i$ represents the $i$-th sentence. We first translate\footnote{https://translate.google.com/} them into a pivot language (e.g., French) and then translate them back into English, as Figure \ref{fig:fst_boost} shows. In this way, we obtain a rewritten sentence $\bm{s}_i'$ for each sentence $\bm{s}_i \in \mathcal{S}$.

To verify whether $\bm{s}_i'$ improves the formality compared to $\bm{s}_i$, we introduce a formality discriminator which in our case is a Convolutional Neural Network (CNN) to quantify the formality level of a sentence. We trained the formality discriminator with the sentences and their formality labels in the FST corpus (e.g., GYAFC). The pairs ($\bm{s}_i$, $\bm{s}_i'$) where $\bm{s}_i'$ largely improves the formality of $\bm{s}_i$ will be selected as the augmented data. The resulting data set $\mathcal{T}_{aug}$ is such a set of pairs:
\begin{equation}\label{eq}
 \mathcal{T}_{aug} = \{(\bm{s}_i, \bm{s}_i') | P_{+}(\bm{s}_i') - P_{+}(\bm{s}_i) \ge \sigma \}
\end{equation}
where $P_{+}(\bm{x})$ is the probability of sentence $\bm{x}$ being formal, predicted by the discriminator, and $\sigma$ is the threshold\footnote{$\sigma=0.6$ in our experiments.} 
for augmented data selection. In this way, we can obtain much helpful parallel data with valuable rewriting knowledge that is not covered by the original parallel data.
\subsubsection{Multi-task transfer}\label{subsubsec:multi-task}
In addition to back translation and formality discrimination that use artificially generated sentence pairs for data augmentation, we introduce multi-task transfer that uses annotated sentence pairs from other seq2seq tasks. We observe that informal texts are usually ungrammatical while formal texts are almost grammatically correct. Therefore, a desirable FST model should possess the ability to detect and rewrite ungrammatical texts, which has been verified by the previous empirical study \cite{ge2019automatic} showing that using a state-of-the-art grammatical error correction (GEC) model to post-process the outputs of an FST model can improve the result. Inspired by this observation, we propose to transfer the knowledge from GEC to FST by leveraging the GEC training data as the augmented parallel data to help improve formality. An example is illustrated in Figure~\ref{example} in which the annotated data for GEC provides knowledge to help the model rewrite the ungrammatical informal sentence.

\subsection{Pre-training with Augmented Data}



In general, massive augmented parallel data can help a seq2seq model to learn contextualized representations, sentence generation and source-target alignments better. When the augmented parallel data is available, previous studies~\cite{DBLP:conf/acl/SennrichHB16,edunov2018understanding,DBLP:journals/mt/KarakantaDG18,DBLP:journals/corr/abs-1808-07512} for seq2seq tasks are inclined to train a seq2seq model with original training data and augmented data simultaneously. However, augmented data is usually noisier and less valuable than original training data. In simultaneous training, the massive augmented data tends to overwhelm the original data and introduce unnecessary and even erroneous editing knowledge, which is undesirable for our task.

To better exploit the augmented data, we propose to first pre-train the model with augmented parallel data and then fine-tune the model with the original training data. In our pre-training \& fine-tuning (PT\&FT) approach, the augmented data is not treated equally to the original data; instead it only serves as prior knowledge that can be updated and even overwritten during the fine-tuning phase. In this way, the model can better learn from the original data without being overwhelmed or distracted by the augmented data. Moreover, separating the augmented and original data into different training phases makes the model become more tolerant to noise in augmented data, which reduces the quality requirement for the augmented data and enables the model to use noisier augmented data and even training data from other tasks.

\section{Experiments}\label{sec:experiment}

In this section, we present the experimental settings and related experimental results. We focus on informal$\to$formal style transfer since it is more practical in real application scenarios.

\subsection{Experimental Settings}
We use GYAFC benchmark dataset \cite{DBLP:conf/naacl/RaoT18} for training and evaluation. GYAFC's training split contains a total of 110K annotated informal-formal parallel sentences, which are annotated via crowd-sourcing of two domains: \emph{Entertainment \& Music} (E\&M) and \emph{Family \& Relationships} (F\&R). In its test split, there are 1,146 and 1,332 informal sentences in E\&M and F\&R domain respectively and each informal sentence has 4 referential formal rewrites. We use all the three data augmentation methods we introduced and obtain a total of 4.9M augmented pairs. Among them, 1.6M are generated by back-translating (BT) formal sentences identified (as formal) by the formality discriminator in E\&M and F\&R domain on Yahoo Answers L6 corpus\footnote{https://webscope.sandbox.yahoo.com/catalog.php}, 1.5M are derived by formality discrimination (F-Dis) by using French, German and Chinese as pivot languages, and 1.8M are from multi-task transfer (M-task) from the public GEC data (Lang-8~\cite{mizumoto2011mining, tajiri2012tense} and NUCLE~\cite{dahlmeier2013building}). The informal sentences used in F-Dis strategy are also from Yahoo Answers L6 corpus.

We use the Transformer (base) \cite{vaswani2017attention} as the seq2seq model with a shared vocabulary of 20K BPE~\citep{DBLP:conf/acl/SennrichHB16a} tokens. We adopt the Adam optimizer to pre-train the model with the augmented parallel data and then fine-tune it with the original parallel data. In pre-training, the dropout rate is set to 0.1 and the learning rate is set to 0.0005 with 8000 warmup steps and scheduled to an inverse square root decay after warmup; while during fine-tuning, the learning rate is set to 0.00025. We pre-train the model for 80k steps and fine-tune the model for a total of 15k steps.
The CNN we use as the formality discriminator has filter sizes of 3, 4, 5 with 100 feature maps. The dropout rate is set to 0.5. It achieves an accuracy of 93.09\% over the GYAFC test set.
\begin{table}[t]
	\centering
	\scalebox{0.9}{
	\begin{tabular}{l|c|c}
		\hline
		\multirow{2}{*}{\textbf{Model}} &\textbf{E\&M} &\textbf{F\&R} \\  
		&\textbf{$BLEU$}   &\textbf{$BLEU$} \\
		\hline
		Original data &69.44   &74.19  \\
		\hline
        Augmented  data &51.83  &55.66 \\
        \hline
        ST &59.93  &63.16 \\
        ST (up-sampling)  &68.43 &73.04  \\
	    ST (down-sampling) &68.54 &73.69 \\
	    \hline
		PT\&FT &\textbf{72.63} &\textbf{77.01} \\
		\hline
	\end{tabular}}
	\caption{The comparison of simultaneous training (ST) and Pre-train \& Fine-tuning (PT\&FT). Down-sampling and up-sampling are for balancing the size of the augmented data and the original data. Specifically, down-sampling samples augmented data, while up-sampling increases the frequency of the original data.}\label{tab:fst_basic_result}
\end{table}

\subsection{Experimental Results}
\subsubsection{Effect of Proposed Approach}
Table~\ref{tab:fst_basic_result} compares the results of the models trained with simultaneous training (ST) and pre-training \& fine-tuning (PT\&FT). ST with the augmented and original data leads to a performance decline, because the noisy augmented data cannot achieve desirable performance by itself and may distract the model from exploiting the original data in simultaneous training. In contrast, PT\&FT only uses the augmented data in the pre-training phase and treats it as the prior knowledge supplementary to the original training data, reducing the negative effects of the augmented data and improving the results.

\begin{table}[t]
    \centering
	\scalebox{0.9}{
		\begin{tabular}{l|c|c}
			\hline
			\multirow{2}{*}{\textbf{Model}} &\textbf{E\&M} &\textbf{F\&R} \\  
			&\textbf{$BLEU$}   &\textbf{$BLEU$}   \\
			\hline
			Original data &69.44   &74.19  \\
            \hline
             \multicolumn{3}{c}{\textbf{Pre-training \& Fine-tuning}} \\
            \hline
		    + BT   &71.18   &75.34   \\
		    + F-Dis     &71.72    &76.24   \\
		    + M-Task    &71.91     &76.21  \\
			+ BT + M-Task + F-Dis &\textbf{72.63}   &\textbf{77.01}   \\
			\hline
		\end{tabular}}
	\caption{The comparison of different data augmentation methods for FST.}\label{tab:fst-aug}
\end{table}

Table \ref{tab:fst-aug} compares the results of different data augmentation methods with PT\&FT. Pre-training with augmented data generated by BT enhances the generalization ability of the model, thus we observe an improvement over the baseline. However, it does not introduce any new informal-to-formal transfer knowledge, leading to the least improvement among the three methods. In contrast, both F-Dis and M-Task introduce abundant transfer knowledge for FST. The augmented data of F-Dis includes various informal$\to$formal rewrite knowledge derived from the MT models, allowing the model to better handle the test instances whose patterns are never seen in the original training data; while M-Task introduces GEC knowledge that helps improve formality in terms of grammar. 



We then combine all these beneficial augmented data for pre-training. As expected, the combination strategy achieves further improvement as shown in Table~\ref{tab:fst-aug} since the it enables the model to take advantage of all the data augmentation methods.

\subsubsection{Comparison with State-of-the-Art Results}

We compare our approach to the following previous approaches in the GYAFC benchmark:
\begin{itemize}
    \item Rule, PBMT, NMT, PBMT-NMT: Rule-based, phrase-based MT, NMT, PBMT-NMT hybrid model \cite{DBLP:conf/naacl/RaoT18}.
    \item NMT-MTL: NMT model with multi-task learning \cite{DBLP:conf/coling/NiuRC18}.
    \item GPT-CAT, GPT-Ensemble: fine-tuned encoder-decoder models~\cite{wang-etal-2019-harnessing} initialized by GPT~\cite{radford2019language}. Specifically, GPT-CAT concatenates the original input sentence and the input sentence preprocessed by rules as input, while GPT-Ensemble is the ensemble of two GPT-based encoder-decoder models: one takes the original input sentence as input, the other takes the preprocssed sentence as input.
\end{itemize}
\begin{table}[t] 
    \centering
	\scalebox{0.82}{
		\begin{tabular}{l|c|c}
			\hline
			\multirow{2}{*}{\textbf{System}} &\textbf{E\&M} &\textbf{F\&R}  \\  
			&\textbf{$BLEU$}  &\textbf{$BLEU$}  \\
		    \hline
			No-edit &50.28 &51.67 \\
			\hline
			
			Rule &60.37 &66.40  \\
			PBMT  &66.88 &72.40  \\
			NMT &58.27 &68.26 \\
			NMT-PBMT  &67.51  &73.78  \\
			NMT-MTL    &71.29 &74.51  \\
			NMT-MTL-Ensemble*~~~~~~~~~ & 72.01  &75.33 \\
			GPT-CAT  &72.70 &77.26 \\
			GPT-Ensemble*  &69.86 &76.32 \\
            \hline
            Our Approach &72.63 &77.01\\
            Our Approach*  &\textbf{74.24} &\textbf{77.97} \\
            \hline
		\end{tabular}}
	\caption{The comparison of our approach to the state-of-the-art results. * denotes the ensemble results. 
	}\label{tab:fst-sota}
\end{table}

Following \citet{DBLP:conf/coling/NiuRC18}, we train 4 independent models with different initializations for ensemble decoding. According to Table~\ref{tab:fst-sota}, our single model performs comparably to the state-of-the-art GPT-based encoder-decoder models (more than 200M parameters) with only 54M parameters. Our ensemble model further advances the state-of-the-art result only with a comparable model size to the GPT-based single model (i.e., GPT-CAT).


We also conduct human evaluation. Following \citet{DBLP:conf/naacl/RaoT18}, we assess the model output on three criteria: \emph{formality}, \emph{fluency} and \emph{meaning preservation}. We compare our baseline model trained with original data, our best performing model and the previous state-of-the-art models (NMT-MTL and GPT-CAT). We randomly sample 300 items and each item includes an input and four outputs that shuffled to anonymize model identities. Two annotators are asked to rate the outputs on a discrete scale of 0 to 2. More details can be found in the appendix. The results are shown in Table~\ref{human_eval} which demonstrates that our model is consistently well rated in human evaluation. 

\begin{table}[t]
\centering
	\scalebox{0.9}{
    \begin{tabular}{l|c|c|c}
    \hline
    Model & Formality  & Fluency & Meaning \\ \hline
    Original data &1.31~  &1.77~~~ &1.80~~ \\
    NMT-MTL &1.34~   &1.78~~~ &\textbf{1.92}* \\ 
    GPT-CAT &1.42~ &1.84*~ &1.90~~ \\ \hline
   \textbf{Ours} &~\textbf{1.45}* &~\textbf{1.85}*$^{\dag}$  &\textbf{1.92}* \\
   \hline
    \end{tabular}}
    \caption{Results of human evaluation of FST. Scores marked with */{$^\dag$} are significantly different from the scores of Original data / NMT-MTL ($p < 0.05$ in significance test).}
    \label{human_eval}
\end{table}


\subsubsection{Analysis of Pivot Languages in Feature Discrimination}
We also conduct an exploratory study of the pivot languages used in formality discrimination. Among the three pivot languages (i.e. French, German and Chinese) in our experiments, it is interesting to observe a significant difference in the sizes of the obtained parallel data given the same source sentences and filter threshold, as shown in Table~\ref{size}. Using Chinese as the pivot language results in the most data, probably due to the fact that Chinese and English belong to different language systems. The formality of original informal English sentences may be lost during translation, which turns out to facilitate the MT system to translate Chinese back into formal English. In contrast, French and German have much in common with English, especially for French in terms of the lexicon \cite{baugh1993history}. The translated sentences are likely to maintain informal sense, which hinders the MT system from generating formal English translations.
\begin{table}[t]
\centering
\scalebox{0.9}{
    \begin{tabular}{l|l|l}
    \hline
    French & German & Chinese   \\ \hline
    300k   &530k   &680k    \\ \hline
    \end{tabular}}
    \caption{The sizes of augmented datasets generated by F-Dis based on different pivot languages.}\label{size}
\end{table}

\begin{table}[t] 
	\begin{center}
	\scalebox{0.9}{
		\begin{tabular}{l|c|c}
			\hline
			\multirow{2}{*}{\textbf{Model}} &\textbf{E\&M} &\textbf{F\&R}  \\  
			\cline{2-3}
			&\textbf{$BLEU$}    &\textbf{$BLEU$}   \\
			\hline
			Original data &69.44   &74.19  \\
		    \hline
		    F-Dis (Fr) &70.09  &74.52\\
		    F-Dis (De) &71.15 &75.18  \\
		    F-Dis (Zh) &70.51 &74.79 \\
			\hline
		\end{tabular}}
	\end{center}
	\caption{Performances of formality discrimination based on different pivot languages: French (Fr), German (De) and Chinese (Zh).}\label{solely train}
\end{table}
We compare the performance with augmented data generated by three pivot languages separately in Table~\ref{solely train}. Manual inspection reveals that a few pairs have the issue of meaning inconsistency in all the three sets, which mainly arises from the translation difficulties caused by omissions and poor grammaticality in informal sentences and the segmentation ambiguity in some pivot languages like Chinese. Among the three languages, the Chinese-based augmented data introduces more noise due to the additional segmentation ambiguity problem but brings fair improvement because of its largest size. In contrast, the German-based augmented data has relatively high quality and a moderate size, leading to the best result in our experiments.

\section{Related Work}

Data augmentation has been much explored for seq2seq tasks like Machine Translation ~\cite{he2016dual,DBLP:conf/acl/FadaeeBM17a,zhang2018joint,DBLP:journals/corr/abs-1804-06189,edunov2018understanding,li2019metamt} and Grammatical Error Correction~\cite{kiyono2019empirical,grundkiewicz2019neural,zhao2019improving,zhou2019improving,ge2018fluency, ge2018reaching, xie2018noising,DBLP:conf/bea/YuanBF16,DBLP:conf/bea/ReiFYB17}.
For text style transfer, however, due to the lack of parallel data, many studies focus on unsupervised approaches~\cite{unsupvised_tst1, unsupvised_tst2, unsupvised_tst3} and there is little related work concerning data augmentation. As a result, most recent work \citep{DBLP:journals/corr/JhamtaniGHN17,
DBLP:conf/coling/XuRDGC12} that models text style transfer as MT suffers from a lack of parallel data for training, which seriously limits the performance of powerful models. To solve this pain point, we propose novel data augmentation methods and study the best way to utilize the augmented data, which not only achieves a success in formality style transfer, but also would be inspiring for other text style transfer tasks.

\section{Conclusion}

In this paper, we propose novel data augmentation methods for formality style transfer. Our proposed data augmentation methods can effectively generate diverse augmented data with various formality style transfer knowledge. The augmented data can significantly help improve the performance when it is used for pre-training the model and leads to the state-of-the-art results in the formality style transfer benchmark dataset.

\section*{Acknowledgements}
We thank all the reviewers for providing the constructive suggestions. This work is partly supported by Beijing Academy of Artificial Intelligence. Xu Sun is the corresponding author of this paper.
\bibliography{acl2020}
\bibliographystyle{acl_natbib}

\appendix

\section{Details of Human Evaluation}
We describe the grading standard of the three criteria we present in the main paper for FST: \emph{formality}, \emph{fluency} and \emph{meaning preservation}. The outputs are rated on a discrete scale of 0 to 2. We hire two annotators who major in Linguistics and have received Bachelor degree.

\noindent
\textbf{Formality} Given the informal source sentence and an output, the annotators are asked to rate the formality of a sentence according to the formality improvement level, regardless of fluency and meaning. If the output shows significant formality improvement over the input, it will be rated 2 points. If the output is just slightly more formal than the input, it will be rated 1 point. If the output shows no improvement in the formality or even decreases the formality, it will be rated 0 point.

\noindent
\textbf{Fluency} Given the outputs, the annotators are asked to evaluate the fluency of each sentence in isolation. A sentence is considered to be \emph{fluent} if \emph{it makes sense and is grammatically correct}. The sentences satisfying the requirements will be rated 2 points. The sentences with minor errors will be rated 1 point. If the errors lead to confusing meaning, we give it 0 point.   

\noindent
\textbf{Meaning preservation} Given the output sentence and the corresponding source sentence, the annotators are asked to estimate how much information is preserved of the output compared to the input sentences. If the output sentence and the input exactly convey the same idea, the corresponding system of the output gets 2 points. If they are mostly equivalent but different in some trivial details, the corresponding system gets 1 point. If the output omits some important details that affect the sentence's meaning, the system will get no credit.

For inter-annotator agreement, we calculate the Pearson correlation coefficient of two annotators over the three criteria. The Pearson correlation over the formality criteria is 0.62. For fluency and meaning preservation, the correlation scores are 0.69 and 0.61, respectively.

\end{document}